\documentclass{svmult}
\usepackage{theorem}
\usepackage{graphicx}
\usepackage{amssymb,amsmath}
\begin{document}
\frenchspacing
\title{Robots That Do Not Avoid Obstacles}
\author{Kyriakos Papadopoulos \and Apostolos Syropoulos}
\authorrunning{K. Papadopoulos and A. Syropoulos}
\institute{K. Papadopoulos \at Department of Mathematics, Kuwait University, PO Box 5969, Safat 13060, Kuwait\\ 
\email{kyriakos.papadopoulos1981@gmail.com} \and A. Syropoulos \at Greek Molecular 
Computing Group, Xanthi, Greece  \email{asyropoulos@yahoo.com}}
\maketitle
\abstract{%
The motion planning problem is a fundamental problem in robotics, so that every autonomous robot should be able to deal with it.
A number of solutions have been proposed and a probabilistic one seems to be quite reasonable. However, here we propose a
more adoptive solution that uses fuzzy set theory and we expose this solution next to a sort survey on the recent theory of soft robots,
for a future qualitative comparison between the two.
}
\section{Introduction.}

According to Latombe~\cite{latombe1991}, ``the ultimate goal of robotics is to create autonomous robots''. Farber~\cite{farber:2006} adds that 
\begin{quote}
\ldots{}such robots should be able to accept high-level description of tasks and execute
them without further human intervention. The input description specifies what should
be done and the robot decides how to do it and performs the task. One expects robots
to have sensors and actuators. 
\end{quote}
Typically, robots should be programmed so to be able to plan collision-free motions for complex 
bodies from some point $A$ to another point $B$ while having a collection of static obstacles in 
between. This task is called {\em motion planning}. Naturally, motion planning is very interesting 
but there are many cases where this is not even desirable. For example, a rover moving on the 
surface of a planet should be able to go above obstacles or to even pass through obstacles.

Dynamical systems are characterized by equations that describe their evolution. A dynamical
system is called {\em linear} when its evolution is a linear {\em process}. A process is
linear when a change in any variable at some initial time produces a change in some variable
at some later time, however, if the initial variable changes $n$ times, then the new variable
will change $n$ times at the later time. In other words, any change propagates without
any alterations. Any system that is not linear is called a {\em nonlinear} dynamical
system~\cite{sastry1999}.
A basic characteristic of these systems is that any change in a variable at some initial moment
leads to a change to some variable at a later time, which is not proportional to the initial
change. For example, the {\em logistic map}~\cite{may1976}
\begin{displaymath}
x_{n+1}=rx_{n}(1-x_n),
\end{displaymath}
where $x_n\in[0,1]$ is the magnitude of population in generation $n$ and $x_{n+1}$ the
magnitude of population at generation $n+1$, is a typical example of an equation
that describes a nonlinear system. In this case, the system is the population of some species and
the dynamics the changes from one generation to another.

Although a robotic system can be either linear or nonlinear, it seems that nonlinear
systems are more interesting in terms of applications. A robotic system is called {\em nonlinear} 
when its control is not nonlinear. In particular, a control system is called nonlinear when it
contains at least one nonlinear component~\cite{slotine1991}. For example, a {\em soft 
robot}~\cite{laschi2016}, that is, a robotic system that consists of several deformable spherical
components, is a nonlinear robotic system~\cite{fei2013}. Unlike (some) rigid robots, a soft 
robot can in general go through or above an obstacle. Consider a robot, rigid or soft, that 
moves on a specific path. Assume that we assign to each obstacle which is on this path a 
penetrability degree. Then, the degree to which the robot will not deviate from its path 
to avoid the obstacle will depend on this degree. If the robot can go through the obstacle or 
above it, then we have a nonlinear system moving on a ``vague'' environment. Thus one can say 
that the motion of a soft robot can be desribed also by using fuzzy ``mathematics'' (i.e., 
a very popular mathematical formulation of vagueness).

The central problem of robotics is how to go from point $A$ to point $B$. As explained above, avoiding
obstacles by deviating from a ``predetermined'' path is the ``classical'' way to solve this
problem. However, this is not an interesting problem for us. We are interested in systems that
can use an extended form of the  motion planning algorithm able to describe robots tat go through
or above obstacles. But first, let us examine what is the ``classical'' motion planning algorithm.

\section{Obstacle Avoiding: an up-to-date mathematical formulation.}
Given a vehicle $V$, a starting point $A$ (usually called an initial configuration) and an ending
point $B$ (called a final configuration), one can form the set $P$ of all paths that $V$ can follow,
starting from $A$ and ending in $B$. Clearly, one can define a number of fuzzy subsets of $P$,
for example, the fuzzy subset of easy paths, the fuzzy subset of smooth paths, etc. Obviously, 
the problem is how to chose a path in order to go from $A$ to $B$. This problem is called the
{\em motion planning problem}~\cite{latombe1991}.

A {\em motion-planing algorithm}~\cite{latombe1991} is a solution to the motion
planning problem. Before giving a formal definition to this problem and to its solution,
we describe these notions intuitively. The main task is to find a path starting at
a point $A$ and ending at point $B$. The path has to avoid {\em collisions} with a known set
of stationary obstacles. At any given moment, a robot moving on this path is on a specific
{\em robot configuration} (i.e., a point of this path). In order to solve this problem
one needs a geometric description of both the vehicle and the space where the vehicle moves.
The {\em configuration} $q$ of a vehicle is a specification of the positions of all vehicle points
relative to a fixed coordinate system. The {\em configuration space} is the space of all possible configurations.

Assume that $W \subset \mathbb{R}^3$ is the configuration space on which the vehicle moves, where
$\mathbb{R}^3$ is the Euclidean space of dimension $3$, and denote by $\mathcal{O}\in W$ the set
of all possible obstacles that the vehicle can meet. Such obstacles will be presented in terms of neighborhoods in $\mathbb{R}^3$. The expression $\mathcal{A}(q)$ is used to denote  that the
vehicle is in configuration $q\in C\subseteq W$. Then,
\begin{align*}
C_{\text{free}} &= \Bigl\{ q\in C\mathbin{\Bigm|} \mathcal{A}(q)\cap\mathcal{O}=\emptyset\Bigr\}\\
C_{\text{obs}}  &= C/C_{\text{free}}.
\end{align*}
Let $q_S$ be the initial configuration and $q_G$ the final configuration. Then,
the {\em motion planning problem} is the process of finding a continuous path 
$p:[0,1]\rightarrow C_{\text{free}}$, where $p(0)=q_S$ and $p(1)=q_G$.

One approaches the motion planning problem using different tools and methodologies
and, thus, there are different solutions to it. For example,
Lozano-P\'erez~\cite{lozano:1987} presented a {\em simple} solution, Ashiru
and Czarnecki~\cite{465825} discussed motion planning using genetic
algorithms and Farber~\cite{farber:2006} presented a {\em probabilistic} solution.
Most of all these approaches assume that the vehicle should always avoid obstacles,
but there has not been a study of cases where the vehicle can pass through (penetrate)
an obstacle.

\subsection{A Mathematical Formulation.}
We will use Farber's~\cite{farber:2006} notation and mathematical description of robot motion 
planning algorithm. For topological notions like path-connected spaces, compact-open
topology, etc., see~\cite{engelking1989}.

Let $X$ be a path-connected topological space and denote by $PX$ the space
of all continuous paths. $PX$ is supplied with the compact open topology. Consider
the map $\pi: PX \to X \times X$, which assigns to a path the pair $(\gamma(0),\gamma(1))$ of
the so-called initial-final configurations. $\pi$ is a fibration in the sense of Serre.

\begin{definition}
A {\em motion planning algorithm} is a section $s: X \times X \to PX$ of
fibration, that is, $\pi \circ s = 1_{X \times X}$.
\end{definition}

One of Farber's research goals was to predict the character of instabilities of the behavior
of the robot, knowing several topological properties of the configuration space, such
as its cohomology algebra. Here we will not concern ourselves with this approach.
We will stick in Farber's declaration that there may exist a better mathematical
notion of a configuration space, describing a partially known topological space, whose
(geometric and topological) properties are being gradually revealed. We believe that
fuzzy set theory is the key tool for this.

Farber introduced four numerical invariants $\mathop{\mathrm{TC}}_i(X)$, $i = 1,2,3,4$, measuring
the complexity of the problem of navigation of a robot configuration space. These
invariants coincide for ``good'' spaces, such us for simplicial polyhedra. We will
now present $\mathop{\mathrm{TC}}_4(X)$, for our purposes, 
since it is linked with random motion planning 
algorithms.

\begin{definition}
A random $n$-valued path $\sigma$, on a path-connected topological space $X$, starting at $A \in X$
and ending at $B \in X$ is given by an ordered sequence of paths $\gamma_1,\cdots,\gamma_n \in PX$
and an ordered sequence of real numbers $p_1,\cdots,p_n \in [0,1]$, such that each 
$\gamma_j : [0,1] \to X$ is a continuous path in $X$ starting at $A = \gamma_j(0)$ and ending at 
$B=\gamma_j(1)$, such that $p_j \ge 0$ and $\Sigma_{i=1}^n \gamma_i =1$.
\end{definition}

The notation $P_nX$, of Farber, refers to the set of all $n$-valued random paths in $X$. This set
is a factor-space of a subspace of the Cartesian product of $n$ copies of $PX \times [0,1]$.

\begin{definition}
$\mathop{\mathrm{TC}}_4(X)$ is defined as the minimal integer $n$, 
such that there exists an $n$-valued random motion planning algorithm $s: X \times X \to P_nX$.
\end{definition}
\begin{remark}
It has been proved that $\mathop{\mathrm{TC}}_{n+1}(X)=\mathop{\mathrm{cat}}(X^n)$, for 
$n\ge1$, where $\mathop{\mathrm{cat}}(X^n)$ is the Lusternik-Schnirelmann 
category~\cite{lupton2013}. Î?hese categories have been used to solve problems in nonlinear
analysis (e.g., see~\cite{browder1965}).
\end{remark}

\subsection{Remarks on this Formulation}
No one can doubt the usefulness of Farber's approach, both in the field of Topology
and in Robotics. The instabilities in the robot motion planning algorithm are
linked to topological invariants and the universe where the robot moves is seen
through the eyes of a topologist who sees configuration spaces.
When it comes to engineering though, an interpretation of the invariant $TC_4(X)$ is
tough. What does it mean for a vehicle to take a random path? Is it better to
talk about a plausible path? Moreover, instead of bypassing obstacles, can we assume
that a robot can go through obstacles? 

In what follows, we describe a {\em fuzzy} motion planing problem and explain
how it can be solved. These ideas are explained practically and we conclude with some
questions and problems related to this approach.

\section{Questioning an Even More Theoretical Approach to Motion Planning Problem.}

Here we ask for the possibility of investigating purely topological properties of robot motion planning algorithms via function spaces, based on the study in~\cite{georgpapmeg} and on the 
results by Farber. Considering a function space ${\mathcal{F}(X,Y)}$, there are several 
topological problems one can study. Knowing topological properties of $X$ (or $Y$), what are 
the topological properties of ${\mathcal{F}(X,Y)}$ and vice versa.

Let $X$ be an arbitrary topological space. Let $PX = {\mathcal{C}}([0,1],X)$ be the function 
space of all continuous paths $\gamma : [0,1] \to X$, supplied with the compact-open topology. 
Let $\pi : PX \to X \times X$ be the map which assigns to a path $\gamma$ the pair 
$(\gamma(0), \gamma(1)) \in X \times X$ of the so-called ``initial-final configurations''. 
Consider the function space ${\mathcal{F}}(PX,X \times X)$. A motion planning algorithm is a 
map $s : X \times X \to PX$, such that $\pi \circ s = 1_{X \times X}$. Consider the function 
space ${\mathcal{F}}^M(X \times X, PX)$, consisting of motion planning algorithms. Notice that 
this is a subspace of the function space ${\mathcal{F}}(X \times X,PX)$.

\subparagraph{\textbf{Question 1}}
Farber questions under what conditions there exist motion planing algorithms which are continuous,
and gives an answer through contractibility. More generally,
add (the minimum number of) topological conditions on the function space 
${\mathcal{C}}(X \times X,PX)$, so that its functions to be motion planning algorithms, 
and thus study topological properties of the function space ${\mathcal{C}}^M(X \times X,PX)$ 
of continuous motion planning algorithms. Here we should remark that we did not recommend $X$ to 
be path-connected (which practically means that one can fully control the system by bringing it 
to an arbitrary state from a given state) as an initial condition.

\subparagraph{\textbf{Question 2}}
Start with a topological space $X$, as the configuration space of a mechanical system, with no
explicit information about its local or global topological properties. Apply Step $0$ to Step $n$ 
of the construction given in~\cite{georgpapmeg}, to the motion planning algorithms space 
${\mathcal{F}}(X \times X, PX)$. Study the possibility for the existence of a minimal integer 
$n$ ``revealing as much as possible topological information about $X$''. This will give a partial
answer to Farber's question on robot motion planning algorithms, on whether there exists a way to
study very complex configuration spaces which are gradually revealing their topological properties.

\subparagraph{\textbf{Question 3}}
Given answers to our Question 1, a further theory can be developed, studying the topological
complexity of tame motion planning algorithms, in the language of function spaces 
(see~\cite{farber:2006})

\subparagraph{\textbf{Question 4}}
If a space $X$ is path-connected, one can ``fully control it'', in a sense that for any two 
fixed points there is a path joining them. One could define a topological space, so that for
any two points $A$ and $B$ there exists a {\em linear ordered topological space} (lots) starting
from $A$ and ending at $B$, and this would generalize path-connected spaces and furthermore 
motion planning algorithms.

Can one achieve this in a different way rather than refining the definition of a continuous 
path $\gamma$, by adding the extra property that the path $\gamma$ should be also order 
preserving (taking in $[0,1]$ the natural order $<$)?

One can consider the space of all such lots on X, say $PX$, mapped to $X \times X$ as a fibration 
$\pi$, and define a section $s : X \times X \to PX$, such that $\pi \circ s = id_{X \times X}$. 
One could then study its Schwartz genus, as a notion of a topological complexity of $X$, and 
link notions of order theory and general topology to algebraic topological ideas.

There will be a problem if one considered an arbitrary lots. Consider for
example the lots consisting of just two points can be mapped into any space

$X$ with two points $A$ and $B$ and that mpa will be a homeomorphic embedding, if
and only if $X$ is $T_1$. One does not want this sort of ``teleporting'' behavior to be possible, 
that perhaps one wants there to be many points linking $A$ to $B$ along what ``resembles a 
path''. A general way to achieve this is to require that the lots to be a dense order. If one 
follows this route, it would be most natural to require paths to be closed subsets and the 
map to be a homeomorphic embedding. Alternatively, one could fix a lots $L$ that is to work
for all pairs of points in the space:
\begin{enumerate}
\item when $L=\{0,1\}$ then we have a $\mathrm{T}_1$ space and
\item when $L=[0,1]$, then we have a path-connected space.
\end{enumerate}
 What if $ Y = \mathbb{Q} \cap [0,1]$? What if $Y$ is the Cantor set $C$? What
 if $Y = \omega+1$. In either cases, the ``interesting'' spaces are going to
 be totally disconnected

\section{Further Topological Remarks}

For a more detailed discussion, see~\cite{farber:2006}. Here
we add a few more questions of topological nature.

Consider a path-connected topological space $X$. A random $n$-valued path $\sigma$,
in $X$, which starts at point $A$ and ends at point $B$, is given from a sequence
of paths $\gamma_1,\gamma_2,\ldots,\gamma_n$ which belong to $PX$ (the space of
all continuous paths on $X$) and a sequence of real numbers $p_1,p_2,\ldots,p_n$
in $[0,1]$, such that every path $\gamma_j : [0,1] \to X$ is continuous, where
$\gamma_j(0) = A$ and $\gamma_j(1)=B$ and also $p_j \ge 0$ and $p_1+p_2+\cdots+p_n=1$.
From the third Axiom of probability theory, one
induces that $\sigma = p_1\gamma_1 + p_2\gamma_2+\cdots+p_n\gamma_n$.

Consider now the map $\pi : P_n X \to X\times X$, where
$P_nX$ denotes the set of all random $n$-valued paths on $X$. An
$n$-valued random algorithm is a map $s: X\times X \to PX$, such
that $\pi \circ s = 1_{X \times X}$.

In other words, if one considers the pair $(A,B)$ in $X \times X$
(input), the output is an ordered probability distribution
$s(A,B) = p_1\gamma_1 + p_2 \gamma_2 + \cdots + p_n\gamma_n$, that is
the algorithm $s$ induces the path $\gamma_j$ with probability $p_j$.

A first question, is which probability distributions
are outputs of such motion planning algorithms. It would be of
a theoretical interest to characterize probability distributions via
motion planning algorithms. What about if the number of
paths is not countable? If one can define such motion
planning algorithm, then what kind of probability distribution
can one expect as an output? This is a good point to pass
into the next section, which is the approach to the motion
planning problem through fuzzy logic.

\section{Obstacle Avoiding: a Fuzzy Logic approach.}

A fuzzy motion planning problem is a problem that asks how a vehicle can move from a point $A$
to a point $B$ by possibly going through/climb over/penetrate and so on, a number of obstacles,
instead of avoiding them. All obstacles, which are represented mathematically by neighborhoods,
are associated with a {\em traversal difficulty degree} that specifies how difficult it is to go
over a specific obstacle. This degree is a number drawn from $[0,1]$ and when it is equal to 
$1$ for a given obstacle $O$, this implies that $O$ is actually not an obstacle. On the other hand, 
a traversal difficulty degree equal to $0$ means that it is impossible to go over $O$, so the 
robot will have to find ways to avoid it.
\begin{definition}
A fuzzy continuous path is a map $p^{\lambda,\ell}:[0,1]\rightarrow C$ that goes over obstacles 
$O_1,\ldots,O_n\in C_{\text{obs}}$, where the traversal difficulty degree of each obstacle $O_i$ 
is $\lambda_i$, has a plausibility degree that equals $\lambda=\min_{i=0} \lambda_i$ and its length 
is $\ell$.
\end{definition}
Clearly, the smaller the value of $\lambda$ is, the less plausible a specific path is.

\begin{figure}
\centering{\includegraphics[scale=1]{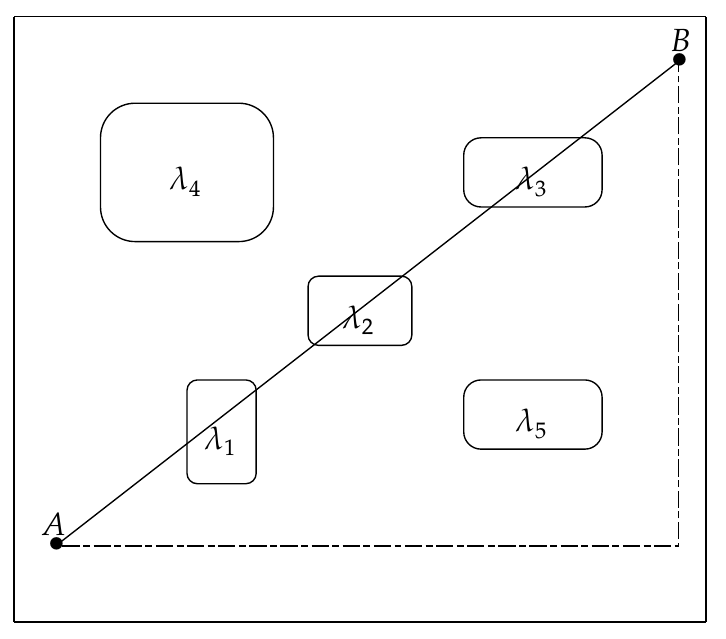}}
\caption{The problem of moving a vehicle from $A$ to $B$ and two possible solutions.}
\label{fuzzypath}
\end{figure}

Figure~\ref{fuzzypath} depicts a terrain with some obstacles. The vehicle's task is to go from 
$A$ to $B$. Obviously, the dotted path is one that avoids all obstacles but it is quite long. 
On the other hand, the straight line is a path that goes over three obstacles but it is the 
shortest possible path. Thus, the ideal path is the one that it will be as short as possible and 
as easy to traverse as possible.

\begin{definition}\label{fuzzy-path}
A {\em fuzzy $n$-valued path} $\sigma$, on $X$, starting at $A \in X$ and ending at $B \in X$ is an
ordered sequence of paths $p^{\lambda_1,\ell_1}_1,p^{\lambda_2,\ell_2}_2,\cdots, p^{\lambda_n,\ell_n}_n \in PX$, where
\begin{displaymath}
\sigma=\min_{\ell_i}\max_{\lambda_i} p^{\lambda_i,\ell_i}_i, \forall i=1,2,\ldots,n.
\end{displaymath}
\end{definition}

Assume that $P_nX$ is the set of all fuzzy $n$-valued paths. Then, the function:
\begin{displaymath}
\pi:P_nX\rightarrow X\times X
\end{displaymath}
maps to a fuzzy path its starting and end points.
\begin{definition}\label{fuzzy-motion-algorithm}
An {\em $n$-valued fuzzy motion planning algorithm} is defined as the map:
\begin{displaymath}
s:X\times X\rightarrow P_nX.
\end{displaymath}
\end{definition}
Thus, the algorithm is a two-fold process: first it identifies $n$ dinstict paths
and it then chooses the most plausible one, not just someone ``in random''.
\begin{remark}
The function $s$ is a continuous section of the fibration $\pi$.
\end{remark}

Having given the above definition of an $n$-valued fuzzy motion planning algorithm, we
now have a clearer picture of how one can define an invariant, similar to 
$\mathop{\mathrm{TC}}_4(X)$
but more realistic, describing its navigational complexity. Let us call such
an invariant $\mathop{\mathrm{TC}}_4^*(X)$. This invariant will depend on both parameters $\lambda$ and $\ell$ of Definition~\ref{fuzzy-path}.
So, it will be sufficient to declare it as the ``smallest integer $n$, such that an $n$-valued
fuzzy motion planning algorithm exists''. $\mathop{\mathrm{TC}}_4^*(X)$ certainly describes
a wider range or properties of the configuration space. Sometimes, in real situations, it will
be better to go through an obstacle, e.g. a vehicle towards water, provided that
in such a way $\ell$ is small, even if $\lambda$ is small too. A mission running
out of time, for example, will put a vehicle into such a risk. In other cases,
it might be better for $\ell$ to be big in order $\lambda$ to be big, too; for instance,
a short distance and a harsh obstacle might put the vehicle into a great risk
or might force it to spend a sufficiently big amount of fuel, etc.

\paragraph{An Example} Imagine that a vehicle, like NASA's Curiosity, is on the surface of planet Mars.
Assume that this vehicle can recognize obstacles and it can assess whether it is possible
to go over an obstacle or not. For example, the rover might have access to an on-board databank
with pictures of obstacles, which have been rated somehow (e.g., by a human expert), and using
some sort of object recognition algorithm, then it can assign traversal difficulty degrees to
various objects and so it can ``deduce'' whether a specific path is traversable or not. More
generally, the vehicle can perform this action several times to find different traversable paths
and to choose the best path. Of course, the system should be able to retract and make another
choice since it is quite possible that some initial estimation was more vague than expected.

\section{Soft Robots or Fuzzy Motion Planning Algorithms?}
On the one hand each obstacle in the path of a robot can be associated with a number that 
will show to what extend it is possible to go through or above the obstacle but on the other
hand we have soft robots that are able to go through obstacles. What is really missing here
is that even for soft robots it would not be absolutely sure that one can go through a
specific obstacle. Thus even for soft robots, each obstacle should be associated with a number
whose value would indicate to what degree it is possible to go through it. In different words,
tbe behavior of soft robots can be better described with the use of fuzzy set theory. Let us
roughly describe how this can be realized.

First we chose the path our robot with follow. Then we assign to each obstacle an
``absolute'' traversal degree, as if our robot is a rigid one. Depending on the shape
of the robot and how flexible it is, we modify the absolute traversal degree so to take
into account the capabilities of the soft robot. The modified traversal degrees can 
be used to define a fuzzy motion planning algorith. The interest thing here is that
the dynamics of the robot are nonlinear and we can use fuzzy sets to desribed a motion
planning algorithm.

\section{Conclusions and Open Questions.}
Αfter describing the motion planning problem problem, we briefly discussed a more
``realistic'' solution and commented on its unsuitability. Next, we presented a formulation 
of the problem that uses ``vagueness'' and proposed a solution that makes use of fuzzy 
set theory. The result is more natural as it coincides with the procedure that humans follow 
in order to choose the most suitable path. We then gave a first comparison of the fuzzy 
formulation with that one that uses soft robots. Here we list a list of open problems which, 
in our own opinion, are interesting both from a theoretical perspective as well as in
applications.

\begin{enumerate}
\item Implement the methodology given in the section ``Obstacle Avoiding: a Fuzzy Logic approach'' with simulation(s) and (an) experiment(s), and see how it works in
practice, comparing it with a similar methodology referring to soft-robots.

\item Nonlinear analysis has been used to analyze fuzzy systems (e.g., see~\cite{jenkins1999}).
Also, tools used to analyze fuzzy systems have been used to analyze nonlinear systems
(e.g., see~\cite{MEI2001759}). The question is: Can use use both methodologies to assist us
to build and test a flexible robot?
\end{enumerate}

\bibliography{robotics_bibl}
\end{document}